%% file: main.tex
\documentclass[runningheads]{llncs}
\usepackage[dvipsnames,table]{xcolor}
\usepackage[year=2026]{accv}
\usepackage[width=122mm,left=12mm,paperwidth=146mm,height=193mm,top=12mm,paperheight=217mm]{geometry}

\usepackage{accvabbrv}
\usepackage[numbers,sort&compress]{natbib} %

\usepackage{graphicx}
\usepackage{booktabs}
\usepackage{tabularx} %

\usepackage[accsupp]{axessibility}  %

\usepackage[pagebackref,breaklinks,colorlinks,citecolor=accvblue]{hyperref}

\usepackage{orcidlink}

\newcommand{\mysection}[1]{\vspace{2pt}\noindent\textbf{#1}}

\begin{document}

\title{SoccerNet 2026 Challenges Results} 

\titlerunning{SN2026}

\author{Anthony Cioppa\inst{1}\textsuperscript{$\star$} \and
Silvio Giancola\inst{2}\textsuperscript{$\star$} \and
Håkan Ardö\inst{3}\textsuperscript{$\dagger$} \and
Mohamad Dalal\inst{4}\textsuperscript{$\dagger$} \and
Jan Held\inst{1,5}\textsuperscript{$\dagger$} \and
Jérémie Ochin\inst{6,7}\textsuperscript{$\dagger$} \and
Jiayuan Rao\inst{8}\textsuperscript{$\dagger$} \and
Karen Sanchez\inst{2}\textsuperscript{$\dagger$} \and
Renaud Vandeghen\inst{1}\textsuperscript{$\dagger$} \and
Artur Xarles\inst{9,10}\textsuperscript{$\dagger$}
% ...
}
\authorrunning{A.~Cioppa et al.}
\institute{$^{1}$\,University of Liège, Liège, Belgium  $\cdot$
$^{2}$\,King Abdullah University of Science and Technology, Thuwal, Saudi Arabia $\cdot$
$^{3}$\,Spiideo, Malmö, Sweden $\cdot$ $^{4}$\,Aalborg University, Aalborg, Denmark $\cdot$
$^{5}$\,SpAItial, Munich, Germany $\cdot$
$^{6}$\,Center for Robotics, Mines Paris, PSL, Paris, France $\cdot$
$^{7}$\,Footovision, Paris, France $\cdot$ $^{8}$\,Shanghai Jiao Tong University, Shanghai, China $\cdot$
$^{9}$\,Universitat de Barcelona, Barcelona, Spain $\cdot$
$^{10}$\,Computer Vision Center, Cerdanyola del Vallès, Spain $\cdot$
$^{11}$\,EVS Broadcast Equipment, Liège, Belgium $\cdot$
$^{12}$\,Pioneer Center for Artificial Intelligence, Copenhagen, Denmark $\cdot$
$^{13}$\,Lund University, Lund, Sweden $\cdot$ $^{14}$\,TAHAKOM, Riyadh, Saudi Arabia $\cdot$
$^{15}$\,Mohamed Bin Zayed University for Artificial Intelligence (MBZUAI), Abu Dhabi, UAE $\cdot$
$^{16}$\,Sber AI, Moscow, Russia $\cdot$ $^{17}$\,Salute For Business, Moscow, Russia $\cdot$
$^{18}$\,Intelligent Perception and Image Understanding Lab, Xidian University, Xi'an, China $\cdot$
$^{19}$\,South China University of Technology, Guangzhou, China $\cdot$
$^{20}$\,Hefei University of Technology, Hefei, China $\cdot$
$^{21}$\,Kyungpook National University, Daegu, Republic of Korea $\cdot$
$^{22}$\,Leipzig University of Applied Sciences, Leipzig, Germany $\cdot$
$^{23}$\,Friedrich-Alexander University Erlangen-Nuremberg, Erlangen, Germany $\cdot$
$^{24}$\,University of Seoul, Seoul, Republic of Korea $\cdot$
$^{25}$\,Shenzhen Institute for Advanced Study, University of Electronic Science and Technology of China, Shenzhen, China $\cdot$
$^{26}$\,Nanjing University, Nanjing, China $\cdot$
$^{27}$\,University of Science, Ho Chi Minh City, Vietnam $\cdot$
$^{28}$\,Nanyang Technological University, Singapore, Singapore $\cdot$
$^{29}$\,National University of Singapore, Singapore, Singapore $\cdot$
$^{30}$\,Chung-Ang University, Seoul, Republic of Korea $\cdot$
$^{31}$\,Artificial Intelligence Society Golem, Warsaw University of Technology, Warsaw, Poland $\cdot$
$^{32}$\,Vietnam National University, Ho Chi Minh City, Vietnam $\cdot$
$^{33}$\,MIXI Inc, Tokyo, Japan $\cdot$ $^{34}$\,AllClip, Inc., Tokyo, Japan $\cdot$
$^{35}$\,University of Tsukuba, Tsukuba, Japan $\cdot$ $^{36}$\,Playbox Inc., Tokyo, Japan $\cdot$
$^{37}$\,Universidad Industrial de Santander, Bucaramanga, Colombia $\cdot$
$^{38}$\,GameChanger by Dick's Sporting Goods, MA, USA $\cdot$
$^{39}$\,Nagoya University, Nagoya, Japan $\cdot$ $^{40}$\,Southeast University, Nanjing, China $\cdot$
$^{41}$\,University of Science and Technology of China, Hefei, China $\cdot$
$^{42}$\,Southwest University, Chongqing, China $\cdot$ $^{43}$\,Peking University, Beijing, China $\cdot$
$^{44}$\,University of North Carolina at Chapel Hill, Chapel Hill, USA $\cdot$
$^{45}$\,Johns Hopkins University, Baltimore, USA $\cdot$
$^{46}$\,Beijing Freedo Technology Co., Ltd., Beijing, China $\cdot$
$^{47}$\,University of California, Berkeley, Berkeley, USA}

\maketitle
\footnotetext{$\star$ Equal contribution. $\dagger$ Task leader. $\ddagger$ Task supervisor.}
\input{sections/0_abstract}

\input{sections/1_introduction}

\input{sections/2_BallActionAnticipation}

\input{sections/3_PlayerCentricBallActionSpotting}

\input{sections/4_NovelViewSynthesis}

\input{sections/5_SpiideoSoccerNetSynloc}

\input{sections/6_VisualQuestionAnswering}

\input{sections/7_Conclusion}


\input{main.bbl}
\include{sections/Appendix}

\end{document}

%% file: sections/0_abstract.tex
\begin{abstract}
  The SoccerNet 2026 Challenges constitute the sixth annual edition of the SoccerNet open benchmarking effort, dedicated to advancing computer vision research in sports video understanding. 
  This year's challenges span five vision-based tasks: (1) \textit{Ball Action Anticipation}, predicting the timing and class of ball-related actions within a short future window from a preceding observation window; (2) \textit{Player-Centric Ball Action Spotting}, temporally localizing and classifying ball-related actions while assigning each action to the acting player through team affiliation and jersey number; (3) \textit{Novel View Synthesis}, rendering images unobserved from camera poses in multi-view football scenes; (4) \textit{Spiideo SoccerNet Synloc}, localizing athletes in real-world pitch coordinates from a single calibrated static-camera image; and (5) \textit{Visual Question Answering}, answering multiple-choice questions about football broadcasts across text, image, and video inputs. 
  For each task, participants were provided with annotated data, a unified evaluation protocol, and a public baseline. This edition saw broad participation, with $427$ teams submitting $1{,}129$ entries across the five tasks and $28$ teams contributing reviewed technical reports. 
  This paper describes each task and its evaluation protocol, presents the challenge leaderboards, and summarizes the leading submissions, with the aim of documenting the current state of each task as measured on held-out challenge data. 
\end{abstract}

%% file: sections/1_introduction.tex
\section{Introduction}
\label{sec:introduction}

\begin{figure*}[t]
    \centering
    \newcommand{\snheight}{2.07cm}
    \begin{tabular}{@{}c@{\hspace{2pt}}c@{\hspace{2pt}}c@{\hspace{2pt}}c@{\hspace{2pt}}c@{}}
        \scriptsize BAA &
        \scriptsize PCBAS &
        \scriptsize NVS &
        \scriptsize SSS &
        \scriptsize VQA \\
        \includegraphics[height=\snheight]{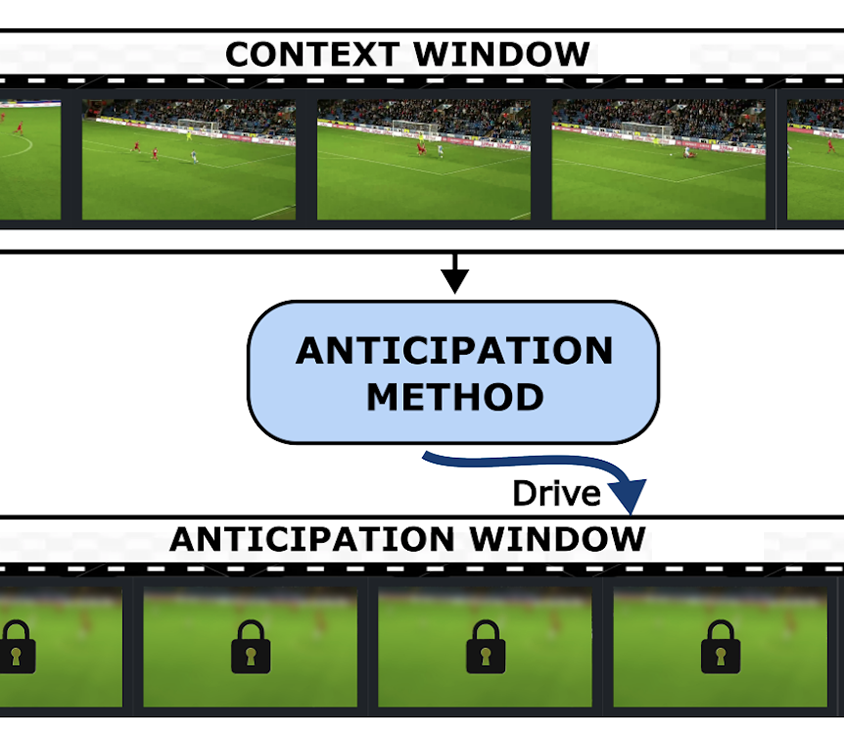} &
        \includegraphics[height=\snheight]{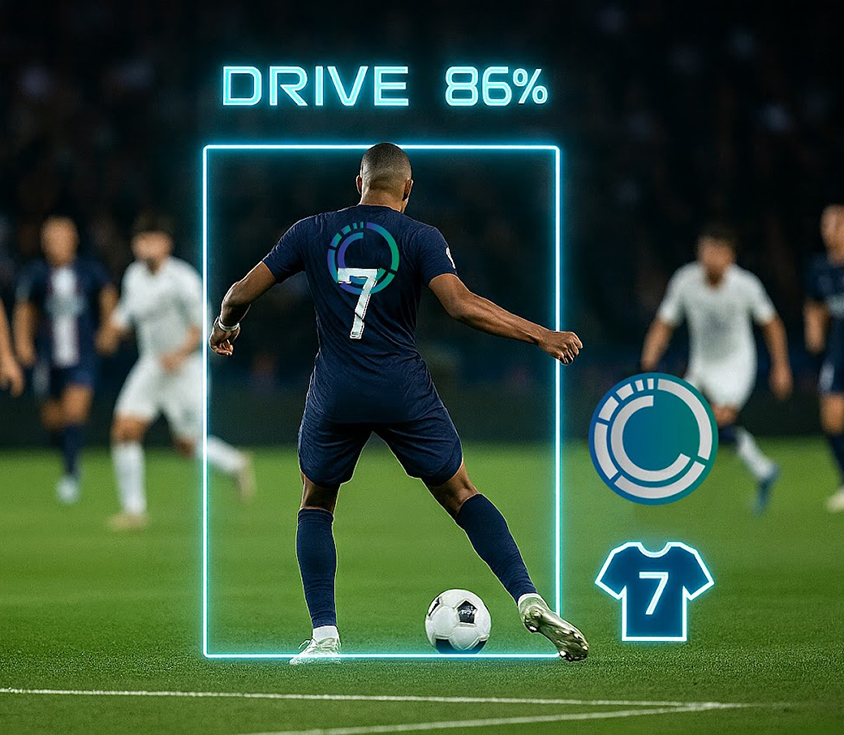} &
        \includegraphics[height=\snheight]{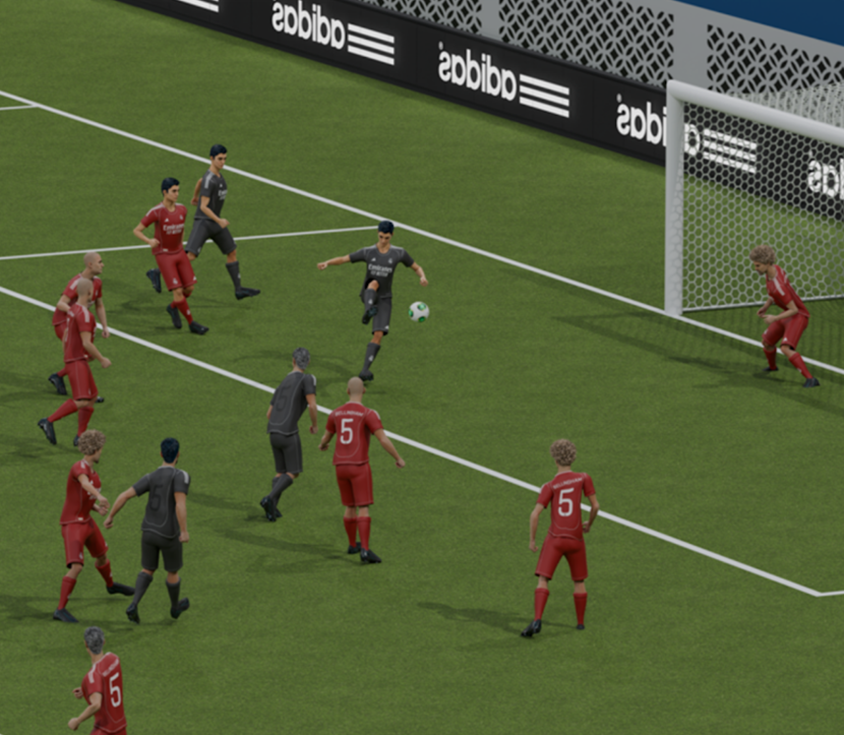} &
        \includegraphics[height=\snheight]{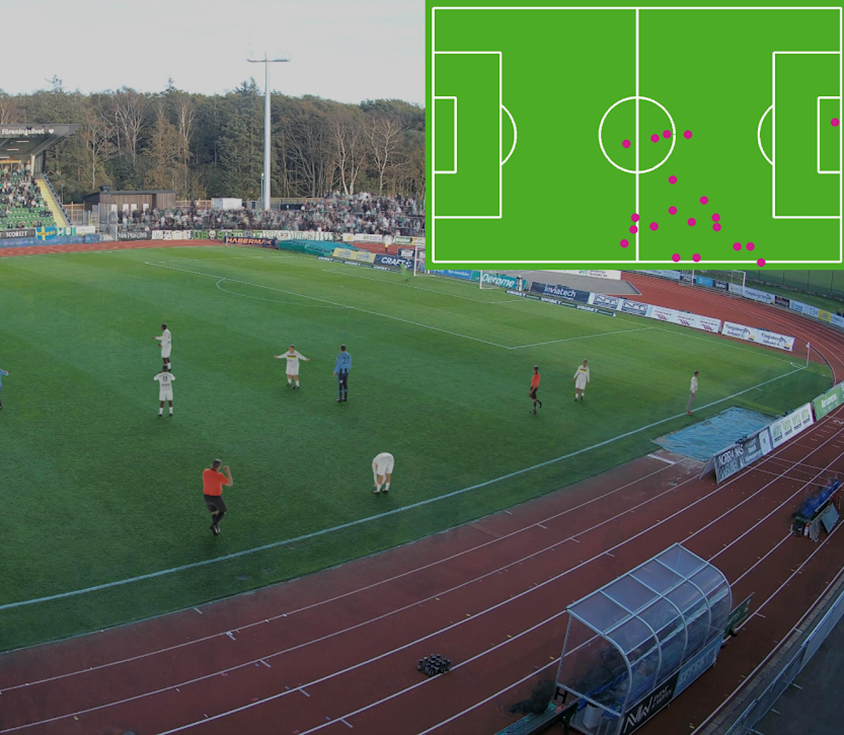} &
        \includegraphics[height=\snheight]{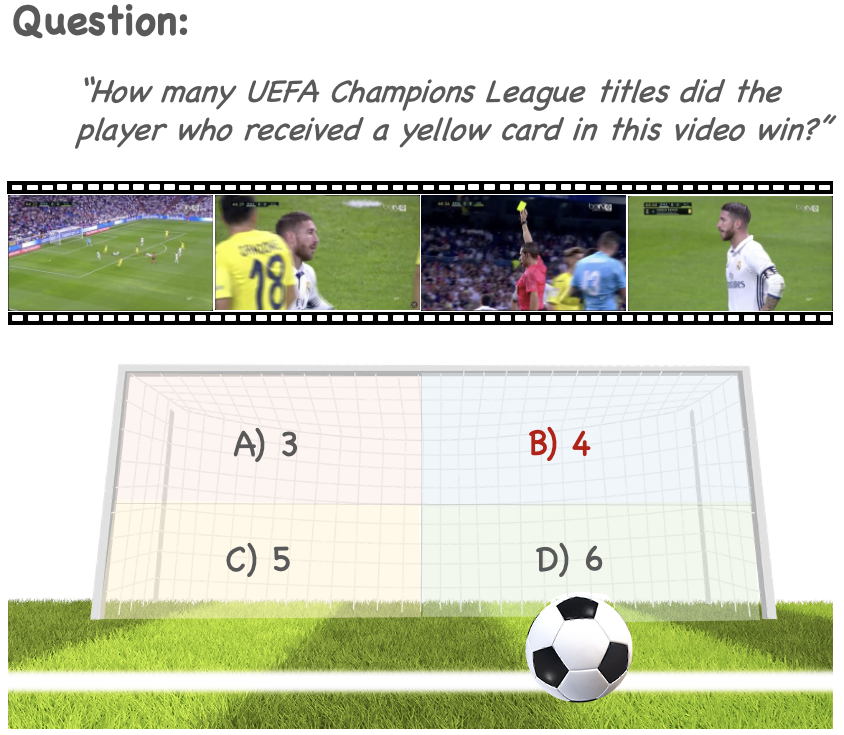} \\
    \end{tabular}
    \caption{\textbf{Overview of the challenges.} In 2026, the SoccerNet challenges encompass five vision tasks:
    (1)~\textit{Ball Action Anticipation} (BAA), focusing on predicting the timing and class of ball-related actions occurring within a five-second window from a preceding 30-second observation window,
    (2)~\textit{Player-Centric Ball Action Spotting} (PCBAS), focusing on temporally localizing and classifying ball-related actions while assigning each action to the acting player through team affiliation and jersey number,
    (3)~\textit{Novel View Synthesis} (NVS), focusing on rendering images from camera poses unobserved during training in multi-view soccer scenes,
    (4)~\textit{Spiideo SoccerNet Synloc} (SSS), focusing on localizing athletes in real-world pitch coordinates from a single calibrated static-camera image, and
    (5)~\textit{Visual Question Answering} (VQA), focusing on answering multiple-choice questions about football broadcasts across text, image, and video.}
    \label{fig:graphical_abstract}
\end{figure*}

Sports video understanding has grown into a well-established area of computer vision research~\cite{Moeslund2014Computer, Thomas2017Computer, Gadde2024TheComputer}, drawing on athlete segmentation~\cite{Cioppa2018ABottomUp, Cioppa2019ARTHuS}, detection~\cite{Vandeghen2022SemiSupervised}, and tracking~\cite{Maglo2022Efficient,Mansourian2023Multitask,Seweryn2025Improving,Somers2024SoccerNetGameState}, re-identification~\cite{Somers2023Body,Mansourian2023Multitask,Balaji2023Jersey}, pose estimation~\cite{Ludwig2025Human, Jiang2024WorldPose}, action recognition~\cite{Held2023VARS,Held2025Towards,Held2024XVARS,Held2025Enhancing,Fang2025Foul, Karki2025Pixels-arxiv} and spotting~\cite{Cioppa2020AContextaware,Cioppa2021Camera,Hong2022Spotting,Soares2022Temporally,Soares2022Action-arxiv,Giancola2023Towards,Seweryn2026Survey,Giancola2025Deep,Cabado2024Beyond,Xarles2024TDEED, Xarles2026AdaSpot}, camera calibration~\cite{Magera2024AUniversal, Magera2025CanGeometry, Theiner2023TVCalib, GutierrezPerez2024NoBells, GutierrezPerez2026PnLCalib, Theiner2026Unified, Falaleev2024Enhancing}, and, increasingly, multimodal and vision-language modeling~\cite{Andrews2024AiCommentator,Gautam2024SoccerNetEchoes, Sarkhoosh2025Beyond, Yang2026SoccerMaster, Rao2024MatchTime, Rao2025MultiAgent, Rao2025Towards}. 
Football is a particularly demanding sport for these computer vision tasks: broadcasts contain many small, visually similar players who frequently occlude one another, the events of interest are brief and sparsely distributed in long untrimmed video, camera viewpoints change abruptly, and the semantics of the game depend on tactical context that is not easily observable from pixels~\cite{Cioppa2022Scaling, Gadde2024TheComputer}. 
Hence, it is a great playground to develop models that must combine fine-grained perception with temporal reasoning.

Progress on such tasks depends on shared data and evaluation protocols that make results comparable across methods. 
The \textit{SoccerNet} initiative was created in 2018~\cite{Giancola2018SoccerNet} to provide this foundation for football video understanding: large-scale annotated datasets released under open licenses for research, public baselines and development kits, and standardized metrics. 
Since 2021, these resources have been accompanied by an annual open challenge~\cite{Giancola2022SoccerNet, Cioppa2024SoccerNet2023Challenge, Cioppa2024SoccerNet2024-arxiv, Giancola2025SoccerNet-arxiv} in which academic and industrial teams submit to a common evaluation server, with rankings determined on a private challenge split. 
The challenge format reduces the risk of fraudulent or non-reproducible evaluation and lowers the barrier to entry for teams without large in-house datasets, while the requirement of technical reports documents the methods behind the leading entries.

\subsection{SoccerNet dataset history and related works}
Giancola~\etal~\cite{Giancola2018SoccerNet} introduced \textit{SoccerNet} in 2018 to provide a large-scale, openly available dataset for reproducible research in football video understanding. 
The original release contained 500 broadcast games drawn from the six major European championships (Serie A, La Liga, Premier League, Ligue 1, Bundesliga, and Champions League), and defined the task of action spotting for the temporal localization of goals, cards, and substitutions. 
The dataset has since been extended through several releases, each adding annotations, modalities, and associated tasks.
Deli\`ege~\etal~\cite{Deliege2021SoccerNetv2} introduced \textit{SoccerNet-v2}, increasing the action spotting annotations to $110{,}458$ actions across $17$ classes and adding annotations for camera shots and replays links.
The first SoccerNet challenge was organized in 2021 around action spotting and replay grounding.
In 2022, Cioppa~\etal~\cite{Cioppa2022Scaling} introduced \textit{SoccerNet-v3} and \textit{SoccerNet-Tracking}~\cite{Cioppa2022SoccerNetTracking}. 
\textit{SoccerNet-v3} added multi-view spatial annotations for players, the ball, field lines, and goal parts, together with three tasks: pitch localization, camera calibration, and player re-identification. 
\textit{SoccerNet-Tracking} introduced multiple object tracking on single-view clips, with metadata such as jersey numbers and team affiliations.
In 2023, \textit{SoccerNet-Captions}~\cite{Mkhallati2023SoccerNetCaption} added natural language descriptions of broadcast events for dense video captioning, and \textit{SoccerNet-MVFouls}~\cite{Held2023VARS} added a multi-view video dataset for foul recognition and characterization.
In 2024, the project released \textit{SoccerNet-Depth}~\cite{Leduc2024SoccerNetDepth} for monocular depth estimation on team-sports video, \textit{SoccerNet-XFoul}~\cite{Held2024XVARS}, a multi-modal dataset of more than $22$k video-question-answer triplets on refereeing decisions,, \textit{SoccerNet-Echoes}~\cite{Gautam2024SoccerNetEchoes} for commentary generation, and \textit{SoccerNet-GSR}~\cite{Somers2024SoccerNetGameState} for game state reconstruction from broadcast video.
More recently, SoccerNet released the action anticipation~\cite{Dalal2025Action}, Spiideo SoccerNet Synloc~\cite{Ardo2025Spiideo}, and group activity recognition~\cite{Karki2025Pixels-arxiv} datasets and tasks, 

The results of the previous editions of the SoccerNet challenges have been reported in dedicated challenge papers~\cite{Giancola2022SoccerNet, Cioppa2024SoccerNet2023Challenge, Cioppa2024SoccerNet2024-arxiv, Giancola2025SoccerNet-arxiv}, which this paper continues for the 2026 edition.
Overall, the methods developed through the SoccerNet challenges have supported many applications such as athlete fatigue assessment~\cite{Bou2026Towards}, game summarization~\cite{Gautam2022Soccer,Midoglu2024AIBased}, pass prediction and feasibility analysis~\cite{ArbuesSanguesa2020Using,Honda2022Pass}, 3D understanding~\cite{GutierrezPerez2025SoccerNetv3D}, and tactics~\cite{Rao2026GenTac-arxiv}. Concurrently, several datasets supported the research community with open-source data such as SoccerDB~\cite{Jiang2020SoccerDB}, SoccerReplay-1988~\cite{Rao2025Towards}, AthleticPose~\cite{Suzuki2025AthleticsPose}, WorldPose~\cite{Jiang2024WorldPose}, or SoccerTrack~\cite{Scott2022SoccerTrack,Scott2025SoccerTrack-arxiv}.

\subsection{SoccerNet 2026 challenges overview}
 
The 2026 edition proposed five vision-based tasks, illustrated in~\cref{fig:graphical_abstract}:
(1) \textit{Ball Action Anticipation}, which requires predicting the timing and class of ball-related actions occurring within a five-second window, given a preceding 30-second observation window, over $10$ action classes~\cite{Dalal2025Action};
(2) \textit{Player-Centric Ball Action Spotting}, which requires temporally localizing and classifying ball-related actions across $8$ classes while also assigning each action to the acting player through team affiliation (left or right) and jersey number~\cite{Ochin2026FOOTPASS};
(3) \textit{Novel View Synthesis}, which requires rendering images from camera poses not observed during training in multi-view soccer scenes;
(4) \textit{Spiideo SoccerNet Synloc}, which requires localizing athletes in real-world pitch coordinates from a single calibrated, static-camera image~\cite{Ardo2025Spiideo}; and
(5) \textit{Visual Question Answering}, which requires answering multiple-choice questions about soccer broadcasts~\cite{Yang2026SoccerMaster}.
Together, these tasks cover complementary forms of perception and reasoning.

Participation in the 2026 edition was substantial. Across the five tasks, $427$ teams submitted $1{,}129$ entries, and $28$ of the top-teams provided reviewed technical reports. %
The leaderboards reported in this paper include only teams that submitted a reviewed technical report. 
This restriction omits some entries for which no report was available, so the reported rankings are not a complete ordering of all submissions. 
In the main paper, we summarize the winning submission for each task, and the supplementary material collects the technical report summaries of the remaining participating teams, organized by task. The remainder of the paper details each task in turn.
 
\mysection{Contributions}. This paper makes the following contributions: (1) it documents the five 2026 tasks and their evaluation protocols, providing a single reference for their definitions, datasets, and evaluation scores, (2), it presents the challenge leaderboards on the held-out challenge splits and summarizes the original contributions of the leading submissions, identifying the design choices that distinguished them, (3) it draws together recurring methodological themes across tasks and notes where performance remains limited, indicating directions where further progress is most needed.

%% file: sections/2_BallActionAnticipation.tex
\section{Ball Action Anticipation}
\label{sec:BAA}

\subsection{Task description}
\label{subsec:BAA_task}

Introduced in~\cite{Dalal2025Action}, Ball Action Anticipation requires models to observe 30 seconds of video and classify and temporally locate ball-related actions occurring in the following 5 seconds. 
The challenge uses the SN-BAA dataset~\cite{Dalal2025Action}, which comprises clips extracted from English Football League matches. The public development data cover 7 matches, while the final challenge evaluation uses 2 additional matches with private annotations. Compared with conventional action-detection settings, BAA is particularly challenging because models must predict unseen events under partial observability and high uncertainty, with the existence of multiple plausible future outcomes.

\subsection{Evaluation scores}
\label{subsec:BAA_eval}

The Ball Action Anticipation task is evaluated using an adaptation of mean Average Precision (mAP) to the anticipation setting~\cite{Dalal2025Action}, at temporal tolerances $\delta \in {1,2,3,4,5,\infty}$. Finite tolerances assess both action recognition and temporal localization, while mAP@$\infty$ evaluates only action recognition. Participants are ranked by $\mathrm{mAP}_{\mathrm{avg}}$, the average of the six mAP@$\delta$ scores, providing a balanced measure of recognition and temporal localization.

\subsection{Leaderboard}
\label{subsec:BAA_leaderboard}

This year, 9 teams participated in the Ball Action Anticipation challenge, submitting a total of 68 entries, with 5 teams also providing a technical report. The best submission achieved a $\mathrm{mAP}_{\mathrm{avg}}$ of 24.08, outperforming the proposed baseline by 7.32 points. The complete leaderboard can be found in \cref{tab:BAA_winner}.

\subsection{Winner}
\label{subsec:BAA_winner}

\input{teams/1_BAA}

\begin{table}[t]
    \centering
    \setlength{\tabcolsep}{3pt}
    \begin{tabular}{@{}llccccccc@{}}
    \toprule
     & & & \multicolumn{6}{c}{mAP@$\delta$ $\uparrow$} \\
    \cmidrule(lr){4-9}
        Rank & Participant &$\boldsymbol{\mathrm{mAP}_{\mathrm{avg}}}$ $\uparrow$ &  $\delta=$1 & 2 & 3 & 4 & 5 & $\infty$ \\
        \midrule
        1 & FAANTRA-WS & \textbf{24.08} & 9.02 & \textbf{18.18} & \textbf{23.72} & \textbf{27.92} & \textbf{30.18} & \textbf{31.78} \\
        2 & alter & 21.36 & 7.54 & 16.35 & 21.57 & 24.80 & 26.21 & 28.14 \\
        3 & FAANTRA-TS & 21.14 & \textbf{9.21} & 16.61 & 21.38 & 23.92 & 25.51 & 27.32 \\
        \rowcolor{gray!15}
        & Baseline~\cite{Dalal2025Action} & 16.76 & 5.70 & 13.00 & 16.30 & 19.18 & 21.02 & 22.86 \\
        4 & Sarthi-GameChanger & 16.49 & 5.48 & 12.17 & 15.79 & 18.65 & 20.93 & 24.38 \\
        5 & VLM-TCF & 15.27 & 5.25 & 10.89 & 15.34 & 17.90 & 19.23 & 20.73 \\
        \bottomrule
    \end{tabular}
    \caption{\textbf{Ball Action Anticipation.} The main evaluation metric, $\mathrm{mAP}_{\mathrm{avg}}$, is highlighted in bold, as are the best results for each metric. We report only teams with submitted and reviewed technical reports, and their summary can be found in the Supplementary Material or \cref{subsec:BAA_winner} for the winning team.}
    \label{tab:BAA_winner}
    \vspace{-0.7cm}
\end{table}

\subsection{Results}
\label{subsec:BAA_results}

The first SoccerNet Ball Action Anticipation challenge extends action detection to the more demanding task of predicting actions within an unobserved future window. This setting introduces partial observability and increased uncertainty, as multiple future action sequences may be plausible from the same observed context. 

Most submissions build on the FAANTRA baseline~\cite{Dalal2025Action}, which uses a standard transformer encoder-decoder with learnable action queries to predict future action classes and timings, trained through sequential query matching and an auxiliary context-segmentation loss. Several methods improved the representation of the observed visual context by increasing the input frame resolution from $224$p to $448$p or $720$p, or by using larger spatial backbones to capture finer-grained visual cues. The resulting richer contextual representations proved beneficial for anticipating actions in the subsequent unobserved interval. Some participants also moved beyond single-model inference by fusing predictions from multiple experts trained with different resolutions or backbone capacities, helping account for the uncertainty inherent in the task and the presence of multiple plausible future outcomes. One method further modeled temporal uncertainty using soft Gaussian targets for action timestamps. Training improvements included a two-stage strategy with increased model capacity in the second stage, exponential moving averages of the model weights to stabilize optimization, and replacing sequential query matching with Hungarian matching to allow greater flexibility in assigning roles to individual queries. VLM-TCF followed the most distinct approach, extracting tactical context with a vision-language model and providing it to FAANTRA for future action anticipation. 

Overall, the challenge results highlight the importance of richer contextual representations, complementary model ensembles, and training strategies that explicitly account for the uncertainty of future actions.

%% file: teams/1_BAA.tex
FAANTRA-WS: Two-Phase Warm-Started FAANTRA with Temporal Calibration for SoccerNet Ball Action Anticipation\\
\textit{Weixuan	Huang (231220091@smail.nju.edu.cn)}\\
FAANTRA-WS adapts the official FAANTRA baseline for SoccerNet Ball Action Anticipation~\cite{Dalal2025Action} using high-resolution temporal modeling and validation-calibrated inference. Training uses $448$p frames, $64$-frame clips, and a $0.5/0.5$ observation-prediction split. The main model uses a RegNetY-GSF backbone and a FUTR-style transformer with eight action queries, trained with a two-phase warm-start recipe: $30$ epochs from ImageNet-pretrained weights followed by $30$ epochs initialized from the full phase-one checkpoint. Class-weighted classification and segmentation losses, $\ell_1$ offset regression, actionness supervision, and a reduced EOS weight improve rare-class performance. At inference, \texttt{RegNetY-008} and \texttt{RegNetY-006} checkpoints are combined through asymmetric logit ensembling, followed by class-wise temporal shifts, and a $10^{-3}$ confidence threshold with empty-clip rescue. The resulting system ranked first place with a challenge $\boldsymbol{\mathrm{mAP}_{\mathrm{avg}}}$ of $24.08$.

%% file: sections/3_PlayerCentricBallActionSpotting.tex
\section{Player-Centric Ball Action Spotting}
\label{sec:PCBAS}

\subsection{Task description}
\label{subsec:PCBAS_task}

Player-centric ball action spotting (PCBAS) extends the spotting tasks introduced in \cite{Cioppa2024SoccerNet2023Challenge, Cioppa2024SoccerNet2024-arxiv, Giancola2025SoccerNet-arxiv}. Participants must identify and temporally localize actions while determining both the team affiliation and jersey number of the acting player. The benchmark considers eight action classes: \textit{Drive}, \textit{Pass}, \textit{Cross}, \textit{Shot}, \textit{Header}, \textit{Throw-in}, \textit{Tackle}, and \textit{Block}, where each action is annotated at a single temporal point corresponding to the moment it occurs. The challenge is built upon the FOOTPASS dataset \cite{Ochin2026FOOTPASS}. All annotated actions are retained, including those occurring during replay segments or close-up shots, making the task particularly challenging under realistic broadcast conditions.

\subsection{Evaluation scores}
\label{subsec:PCBAS_eval}

The task is evaluated using the macro F1-score in a low-confidence, high-recall regime, following \cite{Ochin2026FOOTPASS}. Predictions with confidence below $\tau=0.15$ are discarded. A prediction is considered a true positive if it falls within $\pm12$ frames of a ground-truth annotation and its action class, team affiliation, and jersey number all match the corresponding ground-truth labels. The final score is the average of the class-wise F1-scores over all foreground action classes.

\subsection{Leaderboard}
\label{subsec:PCBAS_leaderboard}

This year, 6 teams participated in the PCBAS challenge, submitting a total of 124 submissions. The best submission achieved a macro F1@0.15 of 58.94, outperforming the proposed baseline, based on Track-Aware Action Detection (TAAD) \cite{Singh2023SpatioTemporal} and Denoising Sequence Transduction Model (DST) \cite{Ochin2025Beyond}, by 12.5 points. The complete leaderboard can be found in \cref{tab:PCBAS_winner}.

\subsection{Winner}
\label{subsec:PCBAS_winner}

\input{teams/1_PCBAS}

{
\setlength{\heavyrulewidth}{1.2pt}
\setlength{\lightrulewidth}{0.6pt}

\begin{table}[t]
    \centering
    \small
    \renewcommand{\arraystretch}{1.05}
    \begin{tabularx}{0.92\columnwidth}{@{}lXr@{}}
    \toprule
        Rank & \hspace{1.2em}Participant & $\boldsymbol{\mathrm{Macro\ F1@0.15}}$ $\uparrow$ \\
    \midrule
        1 & \hspace{1.2em}FSITAHAKOM$^{\mathrm{PCBAS\text{-}1}}$ & \textbf{58.94} \\
        2 & \hspace{1.2em}AISATSANZ$^{\mathrm{PCBAS\text{-}2}}$ & 56.40 \\
        3 & \hspace{1.2em}TeamKIST$^{\mathrm{PCBAS\text{-}3}}$ & 55.69 \\
        4 & \hspace{1.2em}UniBW Munich VIS$^{\mathrm{PCBAS\text{-}4}}$ & 50.35 \\
        \rowcolor{gray!15}
          & \hspace{1.2em}Baseline & 46.41 \\
        5 & \hspace{1.2em}WRF32010$^{\mathrm{PCBAS\text{-}5}}$ & 46.06 \\
        6 & \hspace{1.2em}Sarthi-GameChanger$^{\mathrm{PCBAS\text{-}6}}$ & 44.63 \\
    \bottomrule
    \end{tabularx}
    \caption{\textbf{Player-Centric Ball Action Spotting.} The main evaluation metric, Macro F1@0.15, is highlighted in bold, and higher values indicate better performance. We report only teams with submitted and reviewed technical reports, and their summary can be found in the Supplementary Material or \cref{subsec:PCBAS_winner} for the winning team.}
    \label{tab:PCBAS_winner}
    \vspace{-0.7cm}
\end{table}
}

\subsection{Results}
\label{subsec:PCBAS_results}

Compared to previous SoccerNet action spotting tasks, PCBAS is substantially more challenging because actions must be localized and associated with individual players despite ambiguous visual evidence, occlusions, and off-screen events.

Participating teams explored a wide range of approaches. Several methods improved the baseline components (TAAD \cite{Singh2023SpatioTemporal} and DST \cite{Ochin2025Beyond}) through enhanced temporal modeling, alternative visual backbones, and improved player interaction modeling in the game-state encoder. Other approaches shifted the focus toward game-state reasoning and explicit ball tracking, treating tactical information and player-ball interactions as primary cues while using visual features as complementary information. Some methods also adopted fully player-centric formulations, performing long-range temporal modeling independently for each player track using both visual and tactical features.

Two recurring themes emerged across submissions: ball tracking for extracting ball-centered visual features and player-ball interaction cues, and handcrafted tactical features derived from the provided game-state information. As observed across previous SoccerNet challenges, substantial gains were also obtained through improved training strategies, class-imbalance mitigation, model ensembling, confidence calibration, and dedicated post-processing.

Nevertheless, the \textit{Tackle} class remained particularly challenging for all participants. As the rarest action category in the dataset and one of the most frequently affected by occlusions, it consistently achieved the lowest performance across submissions. These results suggest that further progress will require methods that better handle gaps in player tracklets, maintain player identities through occlusions, and exploit broader game context when visual evidence is ambiguous or entirely missing.

%% file: teams/1_PCBAS.tex
\mysection{PCBAS-1}\\
\text{PAVE - Per-player Attention with agreement-based Voting Ensemble}\\
\textit{Faisal Altawijri and Ismail Mathkour (faltawijri@tahakom.com,
imathkour@tahakom.com)}\\
PAVE is a Per-player Attention and Voting Ensemble method. The main contribution is to strengthen both temporal modeling and model agreement within the TAAD–DST framework. For TAAD, a temporal transformer is added on top of ROI-aligned X3D player features, allowing each player tracklet to be classified with frame-level temporal context. Optimization, scheduling, class weighting, and the number of sampled tracklets are also returned. For DST, the flat role-vector encoding is replaced with per-player attention over the 26 role slots. The strongest variants first model spatial relations between players at each frame, then model each player’s temporal evolution; one parallel spatial-temporal variant is included to increase ensemble diversity. The final PAVE submission combines four independently trained variants with different capacities and attention configurations using weighted event fusion. Agreement filtering suppresses single-model false positives, while a tackle-specific exception preserves recall for the rarest class. PAVE obtains 58.94 Macro-F1 and ranks first.

%% file: sections/4_NovelViewSynthesis.tex
\section{Novel View Synthesis}
\label{sec:NVS}

\subsection{Task description}
\label{subsec:NVS_task}
Novel view synthesis aims to render images of a scene from previously unseen viewpoints, given a set of posed input images.
The task requires reconstructing both the scene geometry and appearance well enough to produce photorealistic views from novel camera poses.
For the task, we created a synthetic dataset with Blender based on real soccer broadcast footage.
We inferred the player position using inverse kinematics based on the 3D joint data derived from real-world player coordinates with manual correction when necessary.
The training set for each scene consists of approximately 420 images rendered at 4K resolution and downsampled versions at lower resolutions.
In addition to the images, we provide the corresponding ground-truth camera parameters in COLMAP format~\cite{Schonberger2016StructurefromMotion}, as well as an initial point cloud inferred from the training views.

\subsection{Evaluation scores}
\label{subsec:NVS_eval}
We evaluate the different methods using the following scores: PSNR, SSIM and LPIPS but only the PSNR is used for ranking following common practice.%
Regarding PSNR, it may favor Gaussian primitives~\cite{Kerbl20233DGaussian,Kheradmand20243DGaussian} rather than sharper primitives~\cite{Held20253DConvex,Held2026Triangle,Held2026MeshSplatting} for complex high-frequency regions, which can represent most of the scene with the grass pitch in close-up views.
In contrast, the SSIM and LPIPS scores are better aligned with human perception for image quality.
In that regard, the best ranking team has been able to improve both baselines of 3DGS~\cite{Kerbl20233DGaussian} and Triangle Splatting~\cite{Held2026Triangle} by more than +3 PSNR, but was unable to improve LPIPS compared to Triangle Splatting.
Compared to traditional benchmarks used in for the task of novel view synthesis~\cite{Barron2022MipNeRF360, Knapitsch2017Tanks} which provide the camera poses and ground-truth images for both training and evaluation, our benchmark only provides the camera poses without the ground-truth images for the evaluation (\textit{challenge}) set.
This approach limits overfitting.

\subsection{Leaderboard}
\label{subsec:NVS_leaderboard}

This year, 65 teams participated in the novel view synthesis challenge, submitting a total of 95 submissions, with 4 teams also providing a technical report. The best submission achieved a PSNR of 29.89, outperforming the proposed baseline by 3.15 points. The complete leaderboard can be found in \cref{tab:NVS_winner}.

\begin{table}[th]
    \centering
    \begin{tabular}{c|ccc}
        Method & \textbf{PSNR} $\uparrow$ & SSIM $\uparrow$ & LPIPS $\downarrow$ \\
        \hline
        Sarthi-GameChanger & \textbf{29.89} & \textbf{0.791} & 0.388 \\
        Ju-Seong Do & 28.94 & 0.785 & 0.364 \\
        Hands-On-Computer-Vision & 28.54 & 0.778 & 0.380 \\
        noual & 28.20 & 0.762 & 0.407 \\
        \hline
        Baseline (3DGS) & 26.74 & 0.751 & 0.410 \\
        Baseline (Triangle Splatting) & 26.43 & 0.757 & \textbf{0.359} \\
    \end{tabular}
    \caption{\textbf{Novel View Synthesis.} The main evaluation score, \emph{PSNR}, is highlighted in \textbf{bold}. Best performances are also shown in \textbf{bold}. Only the teams who provided a reviewed technical reports are reported and their summary can be found in Supplementary Material or \cref{subsec:NVS_winner} for the winning team.}
    \label{tab:NVS_winner}
    \vspace{-0.7cm}
\end{table}

\subsection{Winner}
\label{subsec:NVS_winner}

\input{teams/1_NVS}

\subsection{Results}
\label{subsec:NVS_results}

For this first edition of the Novel View Synthesis challenge, the dataset was designed to make the task accessible to participants with limited prior experience in this field.
Although the training split contains a dense set of viewpoints, the cameras are placed at locations that remain plausible for broadcast production.
The objective is therefore to reconstruct the scene accurately enough to synthesize unseen and interesting viewpoints, which makes novel view synthesis for soccer both appealing and technically challenging.
We provide two baselines based on splatting strategies, 3D Gaussian Splatting (3DGS)~\cite{Kerbl20233DGaussian} and Triangle Splatting~\cite{Held2026Triangle}.
The leaderboard in \cref{tab:NVS_winner} shows that the best submission reaches 29.89 PSNR, improving over the 3DGS baseline by 3.15 dB and over the Triangle Splatting baseline by 3.46 dB, while Triangle Splatting remains the best method according to LPIPS.
The second-ranked submission first densifies the sparse COLMAP point cloud~\cite{Schonberger2016StructurefromMotion} using GaussianPro~\cite{Cheng2024GaussianPro}, before training an antialiased 3DGS model with longer and scene-specific optimization schedules.
The remaining technical reports explore complementary directions, such as combining depth-regularized 3DGS, view-dependent appearance correction, and a heterogeneous ensemble in which Triangle Splatting is used more selectively for players and the ball.
Finally, the report of the last team argues for 2D Gaussian Splatting~\cite{Huang20242DGaussian} as a domain-specific representation for soccer, motivated by the planar structure of the pitch.

%% file: teams/1_NVS.tex
\mysection{NVS-1}\\
DENSER - Depth-Guided Ensemble with Staged EFA-GS Reconstruction for Soccer Novel View Synthesis\\
\textit{Parthsarthi Rawat (sarthi.rawat@gc.com)}\\
\textbf{DENSER} is a \textbf{D}epth-guided \textbf{EN}semble with \textbf{S}taged \textbf{E}FA-GS \textbf{R}econstruction for soccer novel view synthesis. 
The SoccerNet-NVS challenge~\cite{SoccerNet2026NVS} exposes a critical distribution mismatch: ground-level broadcast cameras constitute 59\% of evaluation views yet under 2\% of training cameras, causing standard methods such as 3DGS~\cite{Kerbl20233DGaussian} and Triangle Splatting~\cite{Held2026Triangle} to underfit these views and produce visible artifacts. 
DENSER extends EFA-GS~\cite{Wang2025LowFrequency-arxiv}, an alias-free Gaussian splatting method built on camera-height-based loss weighting that assigns up to $5\times$ weight to ground-level views, (2) scale-and-shift-invariant depth supervision from Depth-Anything-V2~\cite{Yang2024Depth,Yang2024DepthV2} to regularize geometry in textureless regions, and (3) a three-model pixel-average ensemble whose members diverge from a shared 90k-iteration checkpoint by varying training length and Gaussian scale clamping. On five held-out challenge scenes, DENSER achieves a mean PSNR of \textbf{29.89\,dB}, SSIM of \textbf{0.791}, and LPIPS of \textbf{0.366}, ranking \textbf{first} overall.

%% file: sections/5_SpiideoSoccerNetSynloc.tex
\section{Spiideo SoccerNet Synloc}
\label{sec:SSS}

\subsection{Task description}
\label{subsec:SSS_task}
A novel athlete detection and localization task, is introduced, shifting analytics inputs from classical SoccerNet broadcast to static cameras. This task aims to detect and locate athletes on the pitch. Utilizing static cameras eliminates frame-by-frame camera estimation and captures the entire pitch for the entire game. However, it necessitates high-resolution (4K) cameras, leaving distant athletes challenging to handle. Specifically, given an image from a static camera covering half a pitch and its calibration, the objective is to localize each athlete via their pelvis projection onto the ground plane. 

\subsection{Evaluation scores}
\label{subsec:SSS_eval}
The primary evaluation metric is mAP-LocSim \cite{Ardo2025Spiideo}, an average precision metric that uses LocSim to determine correct detections. It is based on the distance $d$ between the predicted and ground-truth positions in real-world pitch coordinates and defined as $e ^ {\ln 0.05 \frac{d^2}{\tau^2}}$, where the distance tolerance $\tau$ is set to $1$ m. This  metric increases with both detection and localization accuracy, but it is difficult to intuitively interpret in absolute terms.
To address that, frame accuracy is also introduced. It evaluates the percentage of images with perfect predictions, that is, zero false positives or negatives and all players correctly detected. A correct detection is defined by a LocSim below 0.5, corresponding to a 0.48-meter distance. Note that evaluation occurs exclusively in pitch coordinates, permitting any image-space athlete representation to be used.

\subsection{Leaderboard}
\label{subsec:SSS_leaderboard}

This year, 88 teams contributed 171 submissions to the Spiideo SoccerNet synloc challenge. The top submission achieved a 97.67 mAP-LocSim, outperforming the baseline by 20 points. \cref{tab:SSS_winner} presents the complete leaderboard.

\begin{table}[t]
    \centering
    \small
    \begin{tabular*}{\columnwidth}{@{\extracolsep{\fill}}clrr}
    \toprule
    Rank & Participant & {\bf mAP-LocSim} & FrameAccuracy \\
    \midrule
        1 &SELabSoccer$^{\mathrm{SSS\text{-}1}}$& {\bf 97.67}& 81.91\\
        2 &PitchSeer$^{\mathrm{SSS\text{-}2}}$&94.95&59.17\\
        3&FC AllClip Research$^{\mathrm{SSS\text{-}3}}$&94.70&{\bf 82.79}\\
        4&Sarthi-GameChanger$^{\mathrm{SSS\text{-}4}}$&94.05&75.92\\
        5&JuMiLe$^{\mathrm{SSS\text{-}5}}$&89.15&71.66\\
        &(Baseline)&77.30&33.74\\
        6&linux godfather$^{\mathrm{SSS\text{-}6}}$&66.92&26.52\\
    \bottomrule
    \end{tabular*}
    \caption{\textbf{Spiideo SoccerNet Synloc.} The main metric (mAP-LocSim) and best performances are bolded. Only teams submitting reviewed technical reports are included. Team summaries are available in the Supplementary Material, with the winning team detailed in \cref{subsec:SSS_winner}.}
    \label{tab:SSS_winner}
\end{table}

\subsection{Winner}
\label{subsec:SSS_winner}

\input{teams/1_SSS}

\subsection{Results}
\label{subsec:SSS_results}

Detecting small athletes required adapting standard techniques to better utilize the high resolution inputs. All teams employed baseline detectors: YOLO26 \cite{Sapkota2025YOLO26-arxiv} (SSS-1, SSS-2, SSS-4, SSS-5), RF-DETR \cite{Robinson2026RF-DETR} (SSS-3), or YOLOX-Pose \cite{Maji2022YOLOPose} (SSS-6). Some utilized auxiliary pose estimators—RTMPose-X \cite{Jiang2023RTMPose-arxiv} (SSS-1) or ViTPose \cite{Xu2024ViTPose++} (SSS-2, SSS-3)—to regress a two-point model (the pelvis and its ground-plane projection). This image-space position was then mapped to the ground plane via the provided camera calibration.

High-resolution processing techniques varied. Tiling was applied adaptively (SSS-1, detailed above) and non-adaptively (SSS-3), with the latter requiring specific NMS techniques to merge duplicate detections of players split across tiles. SSS-4 executed the detector twice—first globally, then on resolution-scaled player crops—enabling subpixel accuracy for distant athletes by upsampling. Finally, SSS-5 retrained their detector at a higher input resolution.

Custom training losses also improved performance. As evaluation occurs in pitch-space, incorporating a world-coordinate Huber Loss (SSS-2) or directly optimizing the LocSim metric (SSS-4) proved beneficial. Standard pixel-based losses suboptimally allocate model capacity, whereas these custom losses prioritize the precise localization of small, distant athletes over large, proximate ones (where equivalent pixel-space errors yield significantly larger real-world coordinate errors). Similar prioritization was achieved via adaptive sample reweighting (SSS-4), assigning higher weights to small or poorly detected players.

Furthermore, the provided camera calibration imposes geometric constraints on the image-space relationship between the pelvis and its ground-plane projection. SSS-5 leveraged this by introducing a deviation-penalizing loss during training and enforcing the constraint during post-processing.

%% file: teams/1_SSS.tex
\mysection{SSS-1}\\
Boundary-Aware Adaptive Tiling and Geometric Keypoint Coupling for Metric  Athlete Localization\\
\textit{Thanh-Khoi Nguyen, 
Hoang-Phuc Nguyen, 
Phuong-Linh Huynh-Ha, and 
Minh-Triet Tran (23120009@student.hcmus.edu.vn, 
nhphuc222@apcs.fitus.edu.vn, 
hhplinh22@apcs.fitus.edu.vn, 
tmtriet@fit.hcmus.edu.vn)}\\
The method is based on a top-down framework for metric-scale athlete localization from single calibrated broadcast frames. The pipeline addresses extreme scale disparities through Boundary-Aware Adaptive Tiling, which dynamically expands fixed-grid crops based on coarse YOLO26-Large detections to guarantee full object containment~\cite{Sapkota2025YOLO26-arxiv, Akyon2022Slicing}. For precise localization, we adapt RTMPose-X~\cite{Jiang2023RTMPose-arxiv} into a two-keypoint estimator predicting only the pelvis and its ground projection. We re-dimension the SimCC head and contract the Gated Attention Unit to operate exclusively on this pair, enforcing their physical coupling under perspective foreshortening. Finally, deterministic ray casting intersects the 2D ground-projection points with the pitch plane, directly lifting predictions into metric world coordinates without learnable height assumptions. Our method achieves state-of-the-art performance, ranking first on the challenge leaderboard with a LocSim score of 97.67 on the private test set.

%% file: sections/6_VisualQuestionAnswering.tex
\section{Visual Question Answering}
\label{sec:VQA}

\subsection{Task description}
\label{subsec:VQA_task}

Visual question answering (VQA) evaluates whether a model can answer natural-language questions from visual evidence~\cite{Antol2015VQA}. The SoccerNet 2026 VQA challenge extends this setting to soccer understanding, where questions may require broadcast perception, temporal localization, match context, and domain knowledge. The task is based on SoccerAgent and SoccerBench~\cite{Rao2025MultiAgent}, with one additional game-state image QA category. Each sample provides a question, four candidate answers, and, when needed, an associated image or video; participants submit one option identifier among \texttt{O1}, \texttt{O2}, \texttt{O3}, and \texttt{O4}.

The benchmark covers $14$ categories across text, image, and video. Text questions focus on background knowledge and match situations. Image questions cover camera views, player identity, jersey numbers, scoreboards, and game-state counting. Video questions cover camera switching, replay grounding, action classification, commentary, jersey color reasoning, and multi-view foul recognition. These tasks reuse SoccerBench/SoccerWiki~\cite{Rao2025MultiAgent}, SoccerReplay-1988~\cite{Rao2025Towards}, MatchTime~\cite{Rao2024MatchTime}, SoccerNet-v2~\cite{Deliege2021SoccerNetv2}, SoccerNet-v3/Jersey Number~\cite{Cioppa2022Scaling}, SoccerNet-Caption~\cite{Mkhallati2023SoccerNetCaption}, SoccerNet-XFoul~\cite{Held2024XVARS}, and SoccerNet Game State Reconstruction~\cite{Somers2024SoccerNetGameState}. The public train and validation splits are derived from SoccerBench, while the hidden test and challenge splits are released through the SoccerNet VQA 2026 benchmark~\cite{SoccerNet2026VQA}.

\subsection{Evaluation scores}
\label{subsec:VQA_eval}

The challenge is evaluated by answer accuracy. For the 500-question challenge split, the score is the percentage of questions for which the submitted option matches the ground truth:
\begin{equation}
    \mathrm{Accuracy} = \frac{\#\mathrm{Correct}}{500} \times 100\%.
\end{equation}

\subsection{Leaderboard}
\label{subsec:VQA_leaderboard}

This year, the visual question answering challenge received 76 submissions, and 8 teams submitted technical reports. The best reported submission achieved an accuracy of 98.0\% on the 500-question challenge split. The leaderboard for teams with submitted reports can be found in \cref{tab:VQA_winner}; a random-choice reference is included for context.

\begin{table}[b]
    \centering
    \small
    \begin{tabular}{clc}
        \toprule
        Rank & Participant & Accuracy (\%) $\uparrow$ \\
        \midrule
        1 & vitomeme & \textbf{98.0} \\
        2 & Sarthi-GameChanger & 96.0 \\
        3 & fkasNeverwinhh & 95.0 \\
        4 & MIXI $\times$ Playbox & 90.0 \\
        11 & vtnhan & 87.0 \\
        12 & nujnow & 86.0 \\
        16 & ysKim & 85.0 \\
        34 & arthur\_g & 59.0 \\
        \rowcolor{gray!15}
        -- & Random & 25.0 \\
        \bottomrule
    \end{tabular}
    \caption{\textbf{Visual Question Answering.} The main evaluation score is accuracy. Best performance is highlighted in bold. Only teams that submitted technical reports are reported.}
    \label{tab:VQA_winner}
\end{table}

\subsection{Winner}
\label{subsec:VQA_winner}

\input{teams/1_VQA}

\subsection{Results}
\label{subsec:VQA_results}

The submitted reports show that the strongest VQA systems did not rely on a single direct prompting strategy. Instead, they routed questions by task type and constructed targeted evidence before selecting an option. The winning team used Gemini as a unified reasoning backbone~\cite{GeminiTeam2023Gemini-arxiv}, while Sarthi-GameChanger built a Claude-based specialist-agent pipeline~\cite{Anthropic2025Sonnet}; fkasNeverwinhh trained a Qwen3-VL-based multi-expert system with LoRA adaptation~\cite{Bai2025Qwen3VL-arxiv,Hu2022LoRA}. Across these approaches, structured context from SoccerAgent/SoccerWiki~\cite{Rao2025MultiAgent}, SoccerReplay-1988~\cite{Rao2025Towards}, and SoccerNet-Caption~\cite{Mkhallati2023SoccerNetCaption} was repeatedly used to turn knowledge-heavy and commentary-related questions into retrieval-grounded reasoning rather than open-ended recall.

The reports also indicate that task-specific visual tools were critical for the harder image and video categories. Several teams refined SoccerAgent-style tool chains by adding person detection and visual prompts~\cite{Redmon2016YOLO}, face or identity matching~\cite{Deng2022ArcFace}, pose-guided jersey-number crops~\cite{Xu2024ViTPose++}, CLIP-style retrieval~\cite{Radford2021Learning}, and player segmentation~\cite{Ravi2025SAM2}. These choices particularly helped with jersey numbers, player/background-knowledge image QA, game-state counting, and replay or foul-related video questions. Remaining errors were concentrated in cases with small or occluded players, ambiguous broadcast viewpoints, noisy entity linking, and temporal evidence that appears only briefly. Overall, the challenge suggests that soccer VQA is best handled by combining strong general VLMs with soccer-specific retrieval and lightweight task-specialized perception modules.

%% file: teams/1_VQA.tex
\mysection{VQA-1}\\
Frontier VLMs are Strong Zero-Shot Soccer Video Reasoners with Proper Elicitation\\
\textit{Xingyu Zhu, 
Yu Zhang, 
Wenwu He, 
Yuyang Sun, 
Haoxuan Ma, 
Yongliang Wu, 
Xiaogang Wang, 
Xinyu Ye, 
Zhenxiang Jiang, 
Yangguang Ji, and 
Wenbo Zhu, 
Xu Yang (xingyu.zhu@nus.edu.sg, 
zhangyu@freedotech.com, 
hewenwu@freedotech.com, 
neilyysun@gmail.com, 
hunterwrynn@gmail.com, 
yongliang0223@gmail.com, 
wangxiaogang@swu.edu.cn, 
xinyuye@cs.unc.edu, 
zhenxiang.jiang@u.nus.edu, 
yji011@e.ntu.edu.sg, 
wenbo\_zhu@berkeley.edu, 
xuyang\_palm@seu.edu.cn)}\\
The method is a task-routed VLM elicitation system for SoccerNet VQA, built on the principle that frontier VLMs already possess strong multimodal perception and reasoning abilities, but require task-specific scaffolding to unlock them reliably. The pipeline routes each question to one of four elicitation paradigms: knowledge grounding, visual prompting, reasoning decomposition, and temporal understanding. For factual questions, structured match metadata, event timelines, web-grounded evidence, and task taxonomies are injected into the context. For spatially demanding tasks, images are augmented with visual prompts such as numbered player boxes and structured person lists. For ambiguous reasoning tasks, observation is separated from judgment through multi-round prompting and database-backed reference anchors. For temporal questions, frame sampling density is adapted according to video duration and  coarse-to-fine analysis is performed when necessary. Using Gemini-3.1-Pro as the unified reasoning engine, the method achieves 97.6\% accuracy on the 500-question test set.

%% file: sections/7_Conclusion.tex
\section{Conclusion}
\label{sec:conclusion}

This paper reported the outcome of the SoccerNet 2026 challenges, the sixth annual edition of the benchmark, covering five vision-based tasks: Ball Action Anticipation, Player-Centric Ball Action Spotting, Novel View Synthesis, Spiideo SoccerNet Synloc, and Visual Question Answering. 
For each task, we described the data and evaluation protocol, reported the leaderboard on the held-out challenge split, and summarized the leading submissions, all of which improved over the provided baselines. 
Recurring gains came from higher input resolution, larger or ensembled models, careful calibration, and the explicit use of domain structure such as camera geometry and tactical features. 
By continuing to release open datasets, standardized protocols, and public baselines, the SoccerNet challenges aim to support reproducible benchmarking in sports video understanding, and we intend to maintain and extend this effort in future editions.

%% file: sections/Appendix.tex
\section{Supplementary Material}
\label{sec:supplementary_material}

\subsection{Ball Action Anticipation}

\input{teams/1_BAA}
\input{teams/2_BAA}

\input{teams/3_BAA}

\input{teams/4_BAA}

\input{teams/5_BAA}

\subsection{Player-Centric Ball Action Spotting}

\input{teams/1_PCBAS}
\input{teams/2_PCBAS}

\input{teams/6_PCBAS}

\input{teams/7_PCBAS}

\subsection{Novel View Synthesis}

\input{teams/1_NVS}
\input{teams/2_NVS}

\input{teams/3_NVS}

\subsection{Spiideo SoccerNet Synloc}

\input{teams/1_SSS}
\input{teams/2_SSS}

\input{teams/3_SSS}

\input{teams/4_SSS}

\input{teams/5_SSS}

\input{teams/6_SSS}

\subsection{Visual Question Answering}

\input{teams/1_VQA}
\input{teams/2_VQA}

\input{teams/3_VQA}

\input{teams/4_VQA}

\input{teams/10_VQA}

\input{teams/11_VQA}

\input{teams/15_VQA}

%% file: teams/2_BAA.tex
\mysection{BAA-2}\\
\text{High-Resolution and EMA-Stabilized Ball Action Anticipation}\\
\textit{Jianling Chu, 
Siyuan Jiang, 
Lechao Cheng, 
Shengeng Tang, 
Yaxiong Wang, and 
Zhun Zhong (2024212135@mail.hfut.edu.cn, 
jsy2585633266@gmail.com, 
chenglc@hfut.edu.cn, 
tangsg@hfut.edu.cn, 
wangyx@hfut.edu.cn, 
zhunzhong007@gmail.com)}\\
Our method builds on the FAANTRA baseline and is trained exclusively on the SoccerNet Ball Action Anticipation dataset, with task-specific refinements tailored to the challenge protocol. To better capture fine-grained ball-centric spatial cues, we increase the input resolution from 224p to 448p while preserving the efficient RegNetY-400MF backbone, Group Shift Fusion module, and query-based transformer encoder-decoder for joint temporal segmentation, action anticipation, and offset regression. To stabilize optimization under the small-batch regime, we maintain an exponential moving average of model parameters with a decay rate of 0.999 and use the EMA model for validation. Moreover, we replace the original tightV2-based checkpoint selection with six-mAP, which averages anticipation mAP across multiple temporal tolerances and is better aligned with the official leaderboard metric. These targeted adaptations yield an average mAP of 21.36.

%% file: teams/3_BAA.tex
\mysection{BAA-3}\\
FAANTRA-TS - FAANTRA-based Two-Stage Training for Ball Action Anticipation in Soccer Broadcasts\\
\textit{Zhenyu Zhao, 
Zihan Zhai, 
Tingting Li, 
Fang Liu,  
Lingling Li, and 
Puhua Chen (25171213940@stu.xidian.edu.cn, 
25171213969@stu.xidian.edu.cn, 
25241215337@stu.xidian.edu.cn, 
f63liu@163.com, 
llli@xidian.edu.cn, 
phchen@xidian.edu.cn)}\\
We propose a FAANTRA-based two-stage training approach for ball action anticipation in soccer broadcasts. Our method leverages the FAANTRA architecture, a Transformer-based model designed to capture long-range temporal dependencies between video frames through self-attention mechanisms. To fully exploit the available data, we adopt a multi-resolution training strategy: the model is first trained on low-resolution (224p) video data for 30 epochs to obtain a robust initialization, followed by fine-tuning on high-resolution (720p) data for 20 epochs to enhance spatiotemporal representation capabilities. Additionally, we implement a careful checkpoint selection strategy to mitigate performance instability caused by validation set fluctuations. Our final submission achieves a score of 21.14 on the 2026 SoccerNet Ball Action Anticipation Challenge test set, significantly outperforming the baseline of 16.76.

%% file: teams/4_BAA.tex
\mysection{BAA-4}\\
Hierarchical GRU with Input-Conditioned Slot Queries for Ball Action Anticipation\\
\textit{Parthsarthi Rawat (sarthi.rawat@gc.com)}\\
We present a hierarchical model for ball action anticipation in football
  broadcast video~\cite{Dalal2025Action}. Given a 30-second observation
  window, the system predicts actions in the subsequent 5-second window
  across 10 action classes. A shared local Transformer encodes clip-level
  features within each 5-second sub-window using a frozen EfficientNetV2
  backbone~\cite{Solovyev2023Ball,Tan2021EfficientNetV2}; a GRU aggregates temporal context
  across all six sub-windows; a Transformer decoder with $K{=}4$
  \emph{input-conditioned} event slots then decodes predictions via three
  decoupled heads (objectness, class, temporal offset). Slot queries are
  seeded from a global GRU summary, enabling adaptation to the specific
  input sequence rather than relying on a fixed initialization. We introduce
  frequency-reweighted Hungarian matching to systematically favor rare
  action classes during training and Gaussian soft targets for smoother
  temporal bin supervision. A weighted random sampler and feature MixUp
  augmentation~\cite{Zhang2018mixup} further mitigate severe class imbalance. On
  the SoccerNet Ball Action Anticipation benchmark~\cite{Dalal2025Action},
  our single-model achieves \textbf{17.91\%} mAP on the test server without
  end-to-end backbone fine-tuning.

%% file: teams/5_BAA.tex
\mysection{BAA-5}\\
\text{VLM-TCF - VLM-Assisted Tactical Context Fusion for Ball Action Anticipation}\\
\textit{Falguni	Ghosh (falguni.ghosh@fau.de)}\\
This work extends the FAANTRA transformer baseline for SoccerNet Ball Action Anticipation by incorporating tactical context extracted using a pretrained vision-language model. A pretrained frozen Qwen2.5-VLM is prompted as a soccer tactical agent to produce compact structured descriptors from sampled video frames, encoding possession, field position, pressure, support, and progression direction. The resulting 20-dimensional context vector is integrated through a dual-pathway fusion mechanism, influencing both encoder representations and decoder queries. This design allows high-level semantic priors to complement learned visual features during anticipation. Under identical training settings, incorporating such structured context leads to improved anticipation performance, particularly under stricter temporal localization constraints. The results suggest that externally derived semantic cues can contribute to action prediction. Future improvements may involve developing soccer-specific VLMs or extracting generic visual context from general VLMs and then using it to build tactical context.

%% file: teams/2_PCBAS.tex
\mysection{PCBAS-2}\\
\text{PC-SSAS - Player-Centric State-Space Action Spotting}\\
\textit{Vadim Linkov, 
Artem Konshin, 
Vasiliy Chelpanov, 
Oleg Durygin, 
Mikhail Moiseev, 
Matvey Isupov, 
Konstantin Mitin, and 
Semen Budennyy (wadim.linkov@gmail.com, 
A.m.konshin@gmail.com, 
vachelpanov@gmail.com, 
oleg.dur97@gmail.com, 
m.moiseev@innopolis.ru, 
matsupus@gmail.com, 
Mitin-uap@yandex.ru, 
Budennyysemen@gmail.com)}\\
We developed a player-centric action spotting system that combines tactical player states, ball detections, and frozen full-frame visual embeddings. Sparse tracking rows are converted into dense frame-player tensors with fixed player slots, kinematic features, visibility masks, ball geometry, possession heuristics, and recent interaction indicators. Actor tokens are initialized from tabular features with role and side embeddings, fused with global visual context through cross-attention, and encoded temporally per player. Frame-level context is obtained by attention pooling over actors and processed with a Mamba state-space temporal backbone~\cite{Gu2024Mamba}, then projected back to player tokens. The model uses eventness, class, and hierarchical branch/fine heads for related action groups. Training uses event-centered sliding windows, Gaussian temporal targets, focal BCE for eventness, and class-balanced sampling. Predictions are calibrated with class-wise threshold tuning and decoded with player-aware suppression.

%% file: teams/6_PCBAS.tex
\mysection{PCBAS-6}\\
\text{Dual-Backbone Fusion and High-Order Geometric Reasoning for PC-BAS}\\
\textit{Ruifeng	Wang, 
Di Yang, and 
Jiangtao Wang (wrf3210@mail.ustc.edu.cn, 
yangdi9860@gmail.com, 
jiangtao.pku@gmail.com)}\\
We propose an enhanced player-centric ball action spotting framework for the SoccerNet 2026 Challenge. Our method refines the baseline in both two stages. First, we employ a dual-backbone visual feature extraction framework combining CNN (X3D-L\cite{Feichtenhofer2020X3D}) and Transformer (Swin3D\cite{Liu2021Swin}) architectures. To resolve temporal jitter and exploit complementary strengths in temporal resolution and spatial context, we propose a decision-level fusion strategy using temporal Gaussian filtering and weighted averaging. Second, we enhance tactical reasoning by introducing High-Order Geometric Features, such as absolute speed and proximity to opponents/teammates, to inject physical priors into the model. Additionally, we integrate a 1D Temporal Convolution Module (TCM) before the Transformer Encoder to provide a local temporal inductive bias, effectively smoothing fragmented logits and suppressing high-frequency noise. Extensive experiments validate that our proposed modules effectively boost Micro F1, precision, and recall.

%% file: teams/7_PCBAS.tex
\mysection{PCBAS-7}\\
\text{Retraining and Post-Processing Extensions to the FOOTPASS Baselines}\\
\textit{Parthsarthi Rawat (sarthi.rawat@gc.com)}\\
We describe our system for the SoccerNet 2026 Player-Centric Ball-Action Spotting Challenge~\cite{Ochin2026FOOTPASS}, which requires jointly predicting \emph{who} performs \emph{which} action and \emph{when} across eight classes in broadcast soccer video.
Building on the three FOOTPASS baselines~\cite{Ochin2026FOOTPASS} ---TAAD~\cite{Singh2023SpatioTemporal,Ochin2025Game}, TAAD+GNN~\cite{Ochin2025Game}, and TAAD+DST~\cite{Ochin2025Beyond}---, we contribute four targeted extensions: (1)~gradient checkpointing to enable full X3D~\cite{Feichtenhofer2020X3D} backbone fine-tuning on a single consumer GPU; (2)~concatenation of GNN logits into the DST encoder, combining graph-based tactical context with per-player visual features to form richer 598-dimensional sequence tokens; (3)~square-root frequency class weighting to address the severe 213:1 pass-to-tackle imbalance in the training corpus; and (4)~a post-processing pipeline comprising per-class logit gating against raw TAAD and GNN scores, temporal frame refinement, jersey re-assignment, and a two-model ensemble. The primary metric is Macro F1, the unweighted mean across all eight action classes. Our system achieves \textbf{0.548} on the test set and \textbf{0.446} on the held-out challenge set.

%% file: teams/2_NVS.tex
\mysection{NVS-2}\\
From Sparse to Dense: GaussianPro Initialization for SoccerNet Novel View Synthesis\\
\textit{Ju-Seong Do, 
Wonyong Jo, 
SuHyun Rim, 
MinJae Kim, 
SeongHeon Kang, and
Ho-Young Jung (jsdo@knu.ac.kr, 
whdnjsdyd111@knu.ac.kr, 
suhyun@knu.ac.kr, 
kmjj139@knu.ac.kr, 
rkd970728@gmail.com, 
hoyjung@knu.ac.kr)}\\
We address the SoccerNet Novel View Synthesis Challenge 2026, which requires rendering novel views of multi-view scenes. Notably, a substantial fraction of challenge views falls outside the spatial coverage of the training cameras, posing a camera extrapolation 
problem rather than interpolation. To address this, we apply GaussianPro's PatchMatch depth propagation~\cite{Cheng2024GaussianPro} to the initial COLMAP cloud before training 3D Gaussian Splatting~\cite{Kerbl20233DGaussian}. The propagation extends depth estimates into regions where the original cloud is sparse, growing it from roughly 200K to between 1.5 and 2.9 million points per scene and yielding a more uniform spatial coverage, including in regions far from training viewpoints. This denser, more uniform initialization is the main factor behind our gains. We also enable the antialiasing filter~\cite{Yu2024MipSplatting}, and our final models reach 28.94 PSNR on the challenge leaderboard (+2.20 dB over the official baseline of 26.74 dB). 
Code: \url{https://github.com/Do-sensei/sn-nvs-2026}.

%% file: teams/3_NVS.tex
\mysection{NVS-3}\\
Pushing the Limits of Novel View Synthesis in Soccer Scenes Through Ensembles\\
\textit{Fabian Perez, 
Juan Vanegas, 
Christian Orduz, and 
Hoover Rueda-Chacón (perez2258059@correo.uis.edu.co, 
juan2221931@correo.uis.edu.co, 
christian.orduz@correo.uis.edu.co, 
hfarueda@uis.edu.co)}\\
Our solution builds upon 3D Gaussian Splatting~\cite{Kerbl20233DGaussian} with the gsplat framework~\cite{Ye2025gsplat}. We improve the baseline through three complementary components: first, we strengthen the appearance module with dropout, layer normalization, and residual color corrections, improving extrapolation to unseen cameras. 
Second, we inject dense monocular depth priors from Depth Anything 3~\cite{Lin2026Depth} and optimize them with a scale-and-shift invariant depth loss plus gradient matching, encouraging consistent geometry. 
Third, we ensemble five independently trained 3DGS models with a Triangle Splatting model~\cite{Held2026Triangle}, using SAM 3 masks~\cite{Carion2026SAM3} to favor sharper player and ball renderings while preserving the background. A final Multi-Variate Gaussian Distribution color transfer~\cite{Hahne2021PlenoptiCam} reduces color drift under extreme viewpoints, yielding stable renderings and consistent gains over the baseline. Code is publicly available at \url{https://github.com/cvail-research/soccernet-nvs-2026}.

%% file: teams/2_SSS.tex
\mysection{SSS-2}\\
\text{PitchSeer}\\
\textit{Mohamed Atef, 
Omar Fetouh, and
Youssef Ghallab \\(mohamed.abouelhadid@mbzuai.ac.ae, 
Omar.elsalakh@mbzuai.ac.ae, 
Youssef.Ghallab@mbzuai.ac.ae)}\\
We present \textit{PitchSeer}, a two-stage top-down framework for single-frame athlete localization in world coordinates using synthetic soccer data. Our method combines YOLO26-pose~\cite{Ultralytics2025YOLO} for athlete detection with ViTPose++~\cite{Xu2024ViTPose++} for fine-grained estimation of pelvis and ground-contact keypoints from player-centered crops. 
The predicted ground points are projected onto the pitch plane using the provided camera calibration parameters to recover metric player locations, following the SynLoc benchmark formulation~\cite{Ardo2025Spiideo}. To better optimize localization performance in the evaluation space, we introduce a projection-aware refinement stage that supervises projected world-coordinate predictions with a Huber-based loss. Experimental results show that higher-resolution 4K inference significantly improves localization accuracy, while the proposed refinement strategy further enhances performance for lightweight models. The proposed pipeline effectively combines accurate detection, transformer-based pose estimation, geometric projection, and projection-space optimization within a compact and efficient framework for athlete localization.

%% file: teams/3_SSS.tex
\mysection{SSS-3}\\
Tile-Consistent Detection and Pitch-Space NMS for Single-Frame Athlete Localization\\
\textit{Ikuma Uchida, 
Minori Sugimura, and 
Takumi Nagaya (ikuma.uchida@allclip.co, 
minorex.0117@gmail.com, 
nagaya.takumi@image.iit.tsukuba.ac.jp)}\\
We present FC AllClip's submission to the Spiideo SoccerNet SynLoc 2026 challenge for single-frame, world-coordinate athlete detection and localization. Our four-stage pipeline performs tile-based player detection with RF-DETR cross-tile fusion via Soft-NMS, top-down pose estimation with a weighted ViTPose ensemble, and a constant pixel shift followed by our pitch-space NMS. 
We make two contributions. First, tile-consistent training and inference align the object-scale distribution between training and deployment, preserving sensitivity to small distant players and lifting detection AP$_{50:95}$ from $0.517$ to $0.878$.
Second, pitch-space NMS projects pelvis-ground keypoints to the bird's-eye view via the ground-truth camera matrix and applies metric-aware greedy suppression, removing duplicates co-located on the pitch yet distant in pixels. 
On the official challenge split, our final model attains mAP-LocSim of $94.70$, a $+17.4$ absolute gain over the baseline, placing third on the final leaderboard.

%% file: teams/4_SSS.tex
\mysection{SSS-4}\\
\text{Two-Stage World-Space Pose Refinement for Precise Soccer Player Localization}\\
\textit{Parthsarthi Rawat (sarthi.rawat@gc.com)}\\
We present a two-stage detection-and-refinement pipeline for sub-meter
soccer player localization in world coordinates from broadcast 4K images.
The first stage employs a YOLO26x~\cite{Sapkota2016VisionBased} pose model operating at 1920\,px resolution on full 4K frames to produce player bounding boxes and coarse ground-projected keypoint estimates.The second stage extracts a padded crop around each detection and applies a second YOLO26x pose model at 640\,px crop resolution to
regress the ground-projected keypoint with sub-pixel precision.
To bridge pixel-space training and metric-space evaluation, we derive
a differentiable coordinate transform ---reversing letterbox scaling,
crop offsets, and perspective camera projection--- and introduce a
\emph{multi-scale LoCSim loss} that jointly penalizes world-space error
at $\tau \in \{0.25, 0.50, 1.0\}$\,m.
The loss is injected only into the one-to-many detection branch,
leaving the one-to-one inference head unaffected while directing
gradients toward small, hard-to-localize players.
On the SpiideoSynLoc~\cite{Ardo2025Spiideo} challenge set, our method achieves
\textbf{94.05\%} mAP-LocSim at $\tau{=}1$\,m and
\textbf{98.90\%} at $\tau{=}5$\,m.

%% file: teams/5_SSS.tex
\mysection{SSS-5}\\
\text{High-Resolution Pose Estimation with Geometric Priors}\\
\textit{Julian Ziegler and Mirco Fuchs (julian.ziegler@htwk-leipzig.de, 
mirco.fuchs@htwk-leipzig.de)}\\
To address the degradation of accuracy for distant subjects in broadcast footage, we implement a two-pronged approach: first, a substantial increase in input spatial resolution to 4k to preserve high-frequency details; second, the integration of geometric priors derived from the camera’s vertical vanishing point. 
This geometric knowledge is employed both as an auxiliary training loss and as a deterministic post-inference correction mechanism to ensure predicted poses align with the scene’s gravitational vector. 
Our empirical results demonstrate that while the high-resolution backbone provides the most significant boost in frame accuracy, the addition of geometric post-processing and Non-Maximum Suppression (NMS) further refines the results and boosts precise localization. This combined approach achieves 89.15\% mAP-LocSim on the challenge set, marking a significant improvement over the baseline.

%% file: teams/6_SSS.tex
\mysection{SSS-6}\\
\text{MS-WBF - Multi-Scale WBF Ensemble for Player Localization}\\
\textit{Jakub Komosa (qbakom@gmail.com)}\\
I adopted the official Spiideo SoccerNet SynLoc baseline~\cite{Ardo2025Spiideo} (YOLOX-Pose Medium with two keypoints per detection: a body anchor and a ground pelvis projected to BEV via the per-image camera matrix). I finetuned the baseline checkpoint for 30 additional epochs on the train split, then ran both checkpoints at three letter-box input sizes (640, 960, 1280) for multi-scale test-time augmentation. The six resulting predictions were combined through a two-level Weighted Box Fusion~\cite{Solovyev2021Weighted} tree at IoU 0.55: per-checkpoint scale fusion, then pairwise checkpoint fusion. The submission score threshold was set to the F1-optimal value computed on the validation split, and all keypoints were rescaled to native 4K coordinates as required by the evaluator. A post-hoc per-camera breakdown revealed that the residual gap to the leading methods is dominated by false-positive over-prediction on high-volume cameras rather than localization error. Final challenge mAP-LocSim: 66.92.

%% file: teams/2_VQA.tex
\mysection{VQA-2}\\
\text{MAESTRO - Multi-Agent Expert System for Task-Routed Operations}\\
\textit{Parthsarthi Rawat (sarthi.rawat@gc.com)}\\
We present \textbf{MAESTRO} (\textbf{M}ulti-\textbf{A}gent \textbf{E}xpert \textbf{S}ystem for \textbf{T}ask-\textbf{R}outed \textbf{O}perations), our solution to the SoccerNet VQA Challenge 2026~\cite{Rao2025MultiAgent}, achieving \textbf{96.0\%} accuracy on the held-out challenge set. 
The challenge presents 500 multiple-choice questions spanning 14 task categories across text, image, and video modalities. MAESTRO is a master-router and specialist-agent pipeline built on Claude Sonnet 4.6~\cite{Anthropic2025Sonnet}. 
A routing agent infers the task category from each question's material path and dispatches to one of 14 specialist sub-agents, each with a dedicated system prompt and context pipeline. 
Two knowledge sources are shared across all agents: SoccerWiki~\cite{SJTU2025SoccerWiki} for player and club entity profiles, and the Game Dataset for match event annotations and timestamps. Metadata-driven context construction maps material paths directly to structured ground-truth records, making inference deterministic for knowledge retrieval tasks. Cost-efficient inference is achieved via the Anthropic Batch API. MAESTRO requires no task-specific fine-tuning, relying entirely on retrieval, structured context injection, and prompt engineering.

%% file: teams/3_VQA.tex
\mysection{VQA-3}\\
\text{MSUE - Multi-Modal Soccer Understanding Expert}\\
\textit{Litao Li, 
Yibo Yu, ,
Yufeng Hu, 
Zhuo Yang, 
Jiali Wen, 
Yixin Chen, and 
Yixi Zhou (seonyee@foxmail.com, 
1635299640@qq.com, 
1797535757@qq.com, 
1115237279@qq.com, 
1583412876@qq.com, 
2762227829@qq.com, 
1140977318@qq.com)}\\
We propose MSUE, a modular multi-expert framework for SoccerNet VQA. MSUE builds 32,151 high-quality instruction samples by converting soccer text, image, and video data into unified caption and MCQ formats, followed by multi-VLM consistency filtering and task-guided dialogue synthesis. The model adopts a two-stage adapted Qwen3-VL-32B backbone, using full-parameter caption tuning for domain visual semantics and LoRA-based VQA tuning for instruction alignment. A Qwen3-4B router dispatches each question to specialized text, image, or video experts, which integrate the fine-tuned visual backbone with SoccerWiki and SoccerReplay retrieval for grounded reasoning. Failure-aware modules further handle jersey color, commentary, and background-knowledge QA. MSUE achieves \textbf{0.95} accuracy on the SoccerNet VQA benchmark, demonstrating the effectiveness of expert routing and retrieval-augmented multimodal adaptation.

%% file: teams/4_VQA.tex
\mysection{VQA-4}\\
\text{Task-Adaptive SoccerAgent for 2026 SoccerNet VQA}\\
\textit{Yuki Nakamura, 
Shun Makino, 
Atom Scott, and 
Rio	Watanabe (yuki.nakamura@playbox.co, 
shun.makino@mixi.co.jp, 
atom@playbox.co, 
rio.watanabe@mixi.co.jp)}\\
Our method builds on SoccerAgent by selectively updating task-specific modules while preserving effective baseline chains. We use Gemini 3.1 Pro Preview as the orchestrator. For image-based QA, we replace the face recognition module with InsightFace to improve player identification. Jersey number recognition combines ViTPose-based body localization, PARSeq OCR, and voting-based aggregation for robust number prediction. For the Game state relevant task, we use RF-DETR and SAM2 to extract player clips, then estimate player roles with an LLM. Video-based tasks are enhanced with speech transcription and time-windowed match-history retrieval, enabling better alignment between visual evidence, commentary, and match context. For visually holistic tasks such as camera status, replay grounding, and multi-view foul recognition, we use direct multimodal inference to avoid unnecessary long chains. This pipeline achieved 90.2\% on CHALLENGE, reaching 4th place in the SoccerNet VQA task. Future work will extend retrieval beyond SoccerWiki and reduce runtime through toolchain pruning.

%% file: teams/10_VQA.tex
\mysection{VQA-10}\\
DREAM - Dynamic Reasoning with Experience-Augmented Memory for SoccerNet 2026-VQA\\
\textit{Thanh-Nhan Vo, 
Thanh-Khoi Nguyen, 
Trong-Thuan Nguyen, 
Trung-Hoang	Le, and 
Minh-Triet Tran (vtnhan@selab.hcmus.edu.vn, 
ntkhoi@selab.hcmus.edu.vn, 
ntthuan@selab.hcmus.edu.vn, 
lthoang@fit.hcmus.edu.vn, 
tmtriet@fit.hcmus.edu.vn)}\\
We propose DREAM, our novel approach to the SoccerNet Challenge 2026-VQA, which redefines the idea of sports multimedia understanding through the integration of structured strategic search and experience-driven learning. DREAM fundamentally treats complex multi-modal queries as state-space exploration using a dynamic DFS-based multi-agent reasoning tree. The framework allows coordinated LLM agents to decompose complex soccer questions into sequential sub-tasks, backtracking and adapting when local reasoning paths fail. To overcome repetitive errors in this search space, we propose an experience-driven episodic memory module. An auxiliary critic analyzes past training failures offline to create succinct correction guidelines that the agent dynamically retrieves during inference. By proactively conditioning the search context, this memory module mitigates hallucinations and effectively guides the DFS path toward an optimal solution. Coupled with majority-voting self-consistency decoding, the framework achieves a remarkable 87\% overall accuracy on the SoccerBench challenge set.

%% file: teams/11_VQA.tex
\mysection{VQA-11}\\
\text{Robust SoccerNet VQA via Task-Specific Tool Refinement}\\
\textit{Wonjun Heo and 
Kwanyong Park (heodnjswns40@uos.ac.kr, 
kwanyong.park@uos.ac.kr)}\\
We improve SoccerAgent with task-specific refinements that reduce open-ended answer generation, long-context matching errors, and dependence on individual samples. For the Camera Detection tool, question options filter few-shot examples, constrain output instructions, and restrict answer parsing, so the model selects only among valid camera labels. For the Replay Grounding tool, we reformulate the task from index generation over long multimodal contexts into retrieval: the replay and each candidate clip are encoded independently using Qwen3-VL-Embedding-8B, and the source clip is selected by cosine similarity. For counting players and referees, our new Grounding Count tool employs task decomposition, replacing direct numeric prediction with color voting, rule-based query parsing, person grounding, and deterministic role and color filtering. For the Face Recognition tool, we aggregate face embeddings from usable reference images and match query images based on minimum distance, reducing dependence on individual frames. These refinements yield a controlled, evidence-based, and robust VQA pipeline.

%% file: teams/15_VQA.tex
\mysection{VQA-15}\\
\text{Task-Adaptive Multimodal Reasoning for SoccerNet Visual Question Answering}\\
\textit{Youngseon Kim and Jongmin Lee
 (thsu1084@cau.ac.kr, 
jmlee@cau.ac.kr)}\\
We present a task-adaptive multimodal reasoning pipeline for SoccerNet Visual Question Answering. Instead of applying a single model uniformly to all questions, our system selects different solving strategies according to the main bottleneck of each task. For background-knowledge questions, it retrieves soccer-domain facts and verifies answers using compact evidence. For image-based knowledge questions, it grounds visual entities with CLIP and FAISS before connecting them to textual records. For localized visual tasks, the pipeline applies task-specific modules such as camera-view classification, jersey-number aggregation, scoreboard cropping, and small-object reasoning. For video and commentary-based tasks, it combines commentary retrieval, rule-based event analysis, dynamic few-shot prompting, and multimodal fallback with stronger models when evidence is weak or ambiguous. This design improves robustness by combining domain knowledge, visual grounding, textual evidence, and conservative fallback decisions across diverse SoccerNet VQA question types.